\documentclass[smallabstract,smallcaptions]{dccpaper}

\usepackage{epsfig}
\usepackage{citesort}
\usepackage{amsmath}
\usepackage{amssymb}
\usepackage{color}
\usepackage{url}
\usepackage{multirow}
\usepackage{booktabs} 
\newlength{\figurewidth}
\newlength{\smallfigurewidth}

\begin{document}

\title
{\large
\textbf{A Neural-network Enhanced Video Coding Framework beyond ECM}
}

\author{%
Yanchen Zhao$^{\ast}$, Wenxuan He$^{\ast}$, Chuanmin Jia$^{\ddag}$, Qizhe Wang$^{\ast}$,\\
Junru Li$^{\dag}$, Yue Li$^{\dag}$, Chaoyi Lin$^{\dag}$, Kai Zhang$^{\dag}$, Li Zhang$^{\dag}$ and Siwei Ma$^{\ast}$\\[0.5em]
{\small\begin{minipage}{\linewidth}\begin{center}
\begin{tabular}{c}
$^{\ast}$School of Computer Science, Peking University\\
$^{\ddag}$Wangxuan Institute of Computer Technology, Peking University\\
$^{\dag}$Multimedia Lab, Bytedance Inc., San Diego CA.92122 USA \\
\url{yczhao@stu.pku.edu.cn, 17wxhe@stu.edu.cn, {cmjia,swma}@pku.edu.cn} \\
\url{{lijunru,yue.li, linchaoyi.cy,zhangkai.video,lizhang.idm}@bytedance.com} \\
\end{tabular}
\end{center}\end{minipage}}
}

\maketitle
\thispagestyle{empty}

\begin{abstract}
In this paper, a hybrid video compression framework is proposed that serves as a demonstrative showcase of deep learning-based approaches extending beyond the confines of traditional coding methodologies. The proposed hybrid framework is founded upon the Enhanced Compression Model (ECM), which is a further enhancement of the Versatile Video Coding (VVC) standard. We have augmented the latest ECM reference software with well-designed coding techniques, including block partitioning, deep learning-based loop filter, and the activation of block importance mapping (BIM) which was integrated but previously inactive within ECM, further enhancing coding performance. Compared with ECM-10.0, our method achieves 6.26\%, 13.33\%, and 12.33\% BD-rate savings for the Y, U, and V components under random access (RA) configuration, respectively. 
\end{abstract}

\section*{Introduction}

In recent years, the application scenarios of video data have undergone continual expansion and enrichment. The rapid increase of video data has posed fresh challenges to the field of image and video compression. Over the past few decades, video coding techniques have been under constant scrutiny in the pursuit of efficiently representing visual signals while maintaining an acceptable perceived distortion. Presently, the prevailing video coding standards, namely High-Efficiency Video Coding (HEVC) \cite{HEVCoverview}, Versatile Video Coding (VVC) \cite{VVCoverview}, and the third generation of the Audio and Video Coding Standard (AVS3) \cite{Zhang2019RecentDO}, all adhere to the classical hybrid coding framework. These frameworks integrate prediction coding \cite{gao2020advanced}, transformation and quantization coding \cite{quantization2019,zhao2018joint}, entropy coding \cite{sze2012entropy}, as well as in-loop filtering \cite{Tsai2013Adaptive} to effectively eliminate spatial, temporal, and statistical redundancies. The most recent iterations of video coding standards, VVC and AVS3, have placed a heightened emphasis on enhancing the adaptability and versatility of video codecs across diverse application scenarios. Notably, both VVC and AVS3 employ a more intricate code unit (CU) partitioning structure to accommodate multi-channel video content better. VVC, for instance, employs a quadtree (QT) nested binary tree and trintree/extended quadtree partitioning \cite{wang2019extended}, while AVS3 even introduces octree block partitioning \cite{avs8k} for ultra-high-definition video. Various techniques used to enhance prediction patterns and enhance motion are also implemented in VVC \cite{VVCoverview} and AVS \cite{yu2009overview}. For instance, methods like cross-component prediction, aimed at eliminating redundant information in different color components, have been applied \cite{li2021crosscomponet}. Additionally, affine motion compensation techniques have been employed to handle non-linear motion \cite{zhang2022advanced}. These approaches contribute to the improvement of prediction accuracy and motion representation in both VVC and AVS.

Due to the swift progress of deep learning in the domain of image and video processing, there has been a growing enthusiasm for exploring deep learning-based tools in hybrid coding frameworks, mainly focusing on intra-prediction, inter-prediction, and filtering. These deep learning-based encoding tools have demonstrated remarkable enhancements in compression performance and reconstruction quality, which play a pivotal role in pushing the boundaries of performance in the exploration experiments of intelligent video coding standards. Various tools have proven effective in integrating with video codecs, including neural-based intra-prediction \cite{intra,dumas2019context}, deep reference frame techniques \cite{zhao2019vrf}, loop filtering \cite{jia2019content,zhao2022cnnlf,10.1145/3502723,10.1145/3529107}, and frame super-resolution \cite{super-resolution}. 


In this paper, we introduce a hybrid video coding framework that effectively combines conventional coding tools with deep learning-based techniques, resulting in a substantial enhancement in compression performance. To be specific, we leverage ECM10.0 \cite{ecm} as a base codec, incorporating an advanced CU-level partitioning structure and deep learning-based filtering while making adjustments to existing tools. Our proposed framework demonstrates a remarkable improvement. Furthermore, the introduction of this hybrid framework is expected to foster increased collaboration and exploration in the convergence of learning-based video compression methods and traditional video compression. 


\begin{figure}[t]
\begin{center}
\begin{tabular}{cccc}
\epsfig{width=1.3in,file=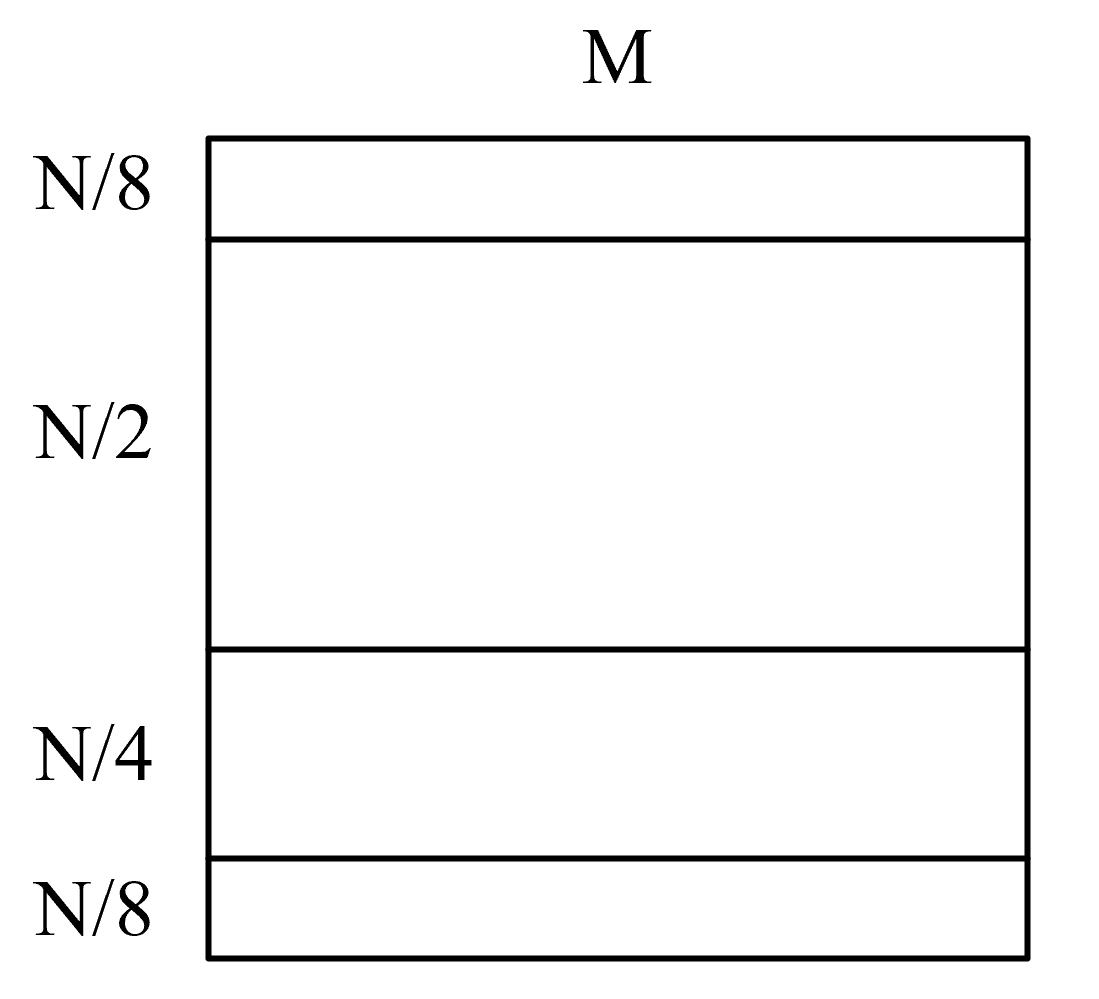} & \epsfig{width=1.3in,file=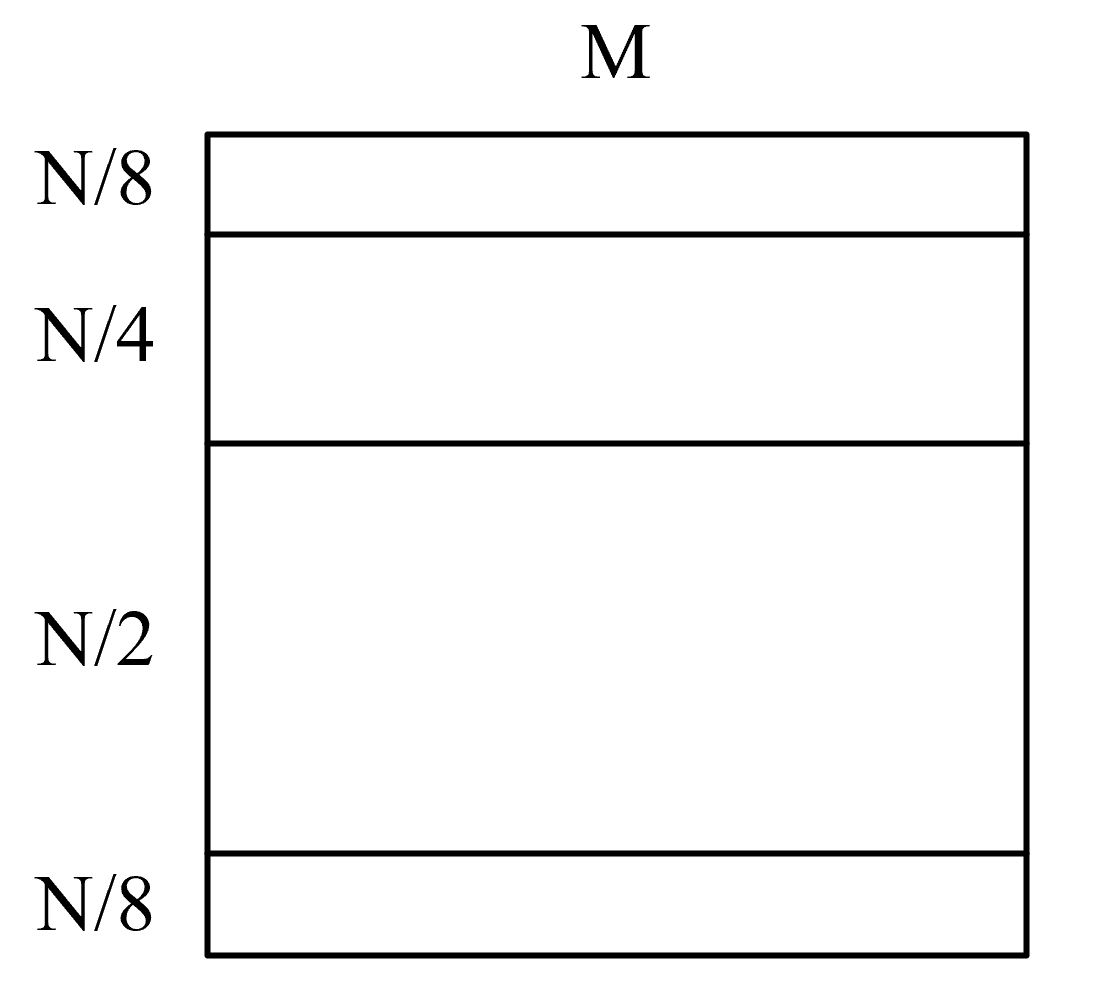} & \epsfig{width=1.3in,file=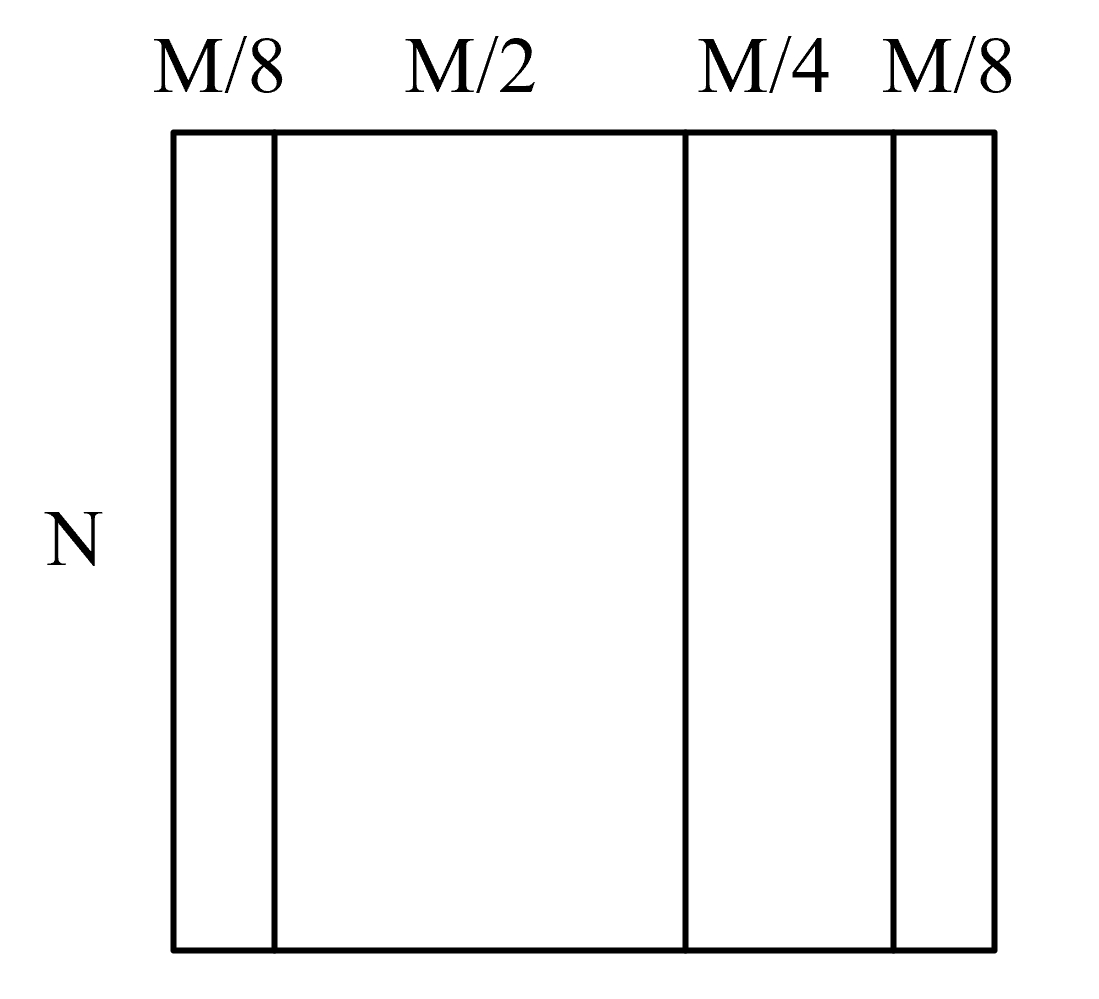} & \epsfig{width=1.3in,file=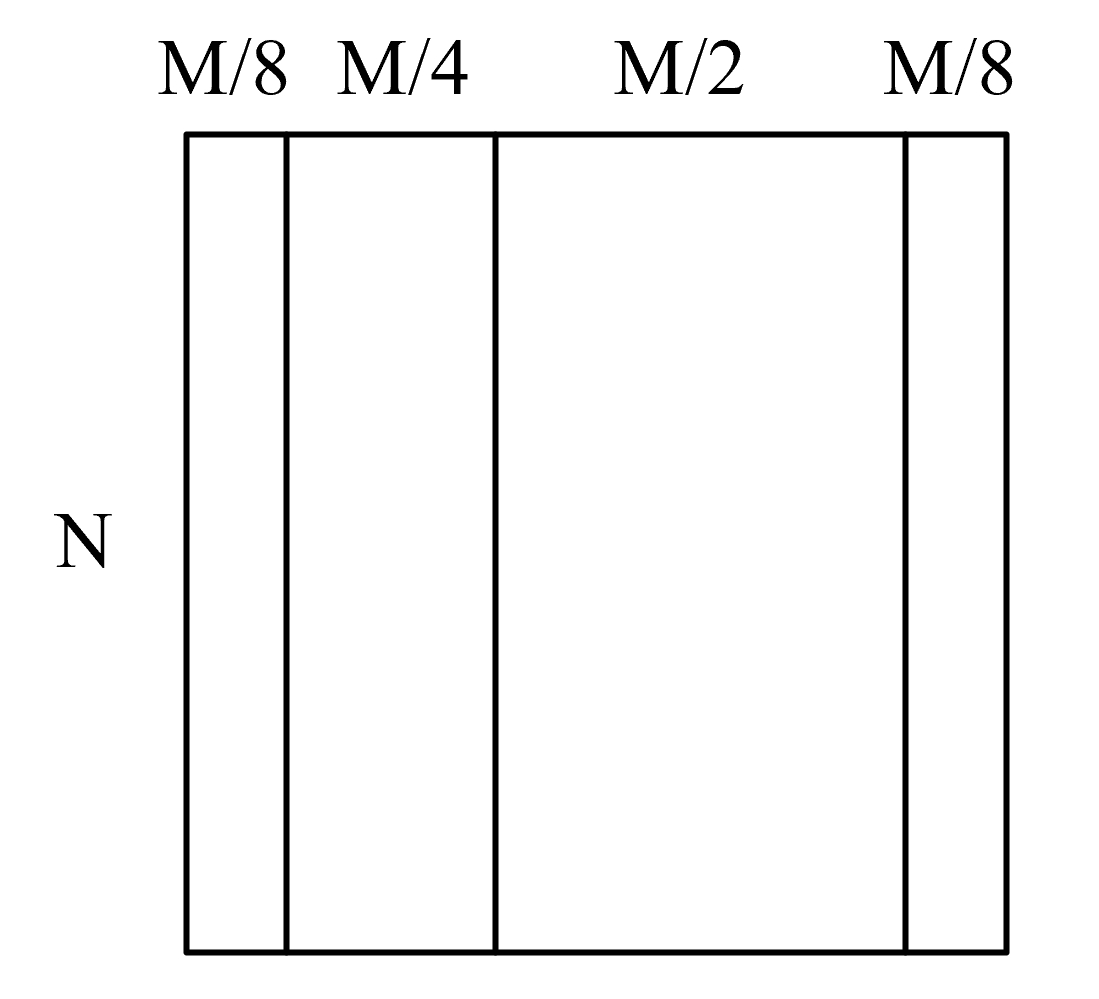} \\
{\small{(a) UQT-H1.}} & {\small(b) UQT-H2.} & {\small(c) UQT-V1.} &{\small(d) UQT-V2.}
\end{tabular}
\end{center}
\caption{\label{fig:uqt}%
Illustration of the UQT partitioning.}
\end{figure}

\section*{Framework}
In this section, we provide a detailed explanation of the elements comprising our video compression framework, which is built on top of the ECM-10.0. It incorporates a set of effective traditional coding tools. We seamlessly integrated asymmetric quaternary tree technology and neural network-based video coding technology into this hybrid coding framework and attempted to enable block importance mapping technology. The resulting learning-based framework attains impressive compression performance through the synergy of advanced coding techniques.

\begin{figure}[t]
\begin{center}
\epsfig{width=5in,file=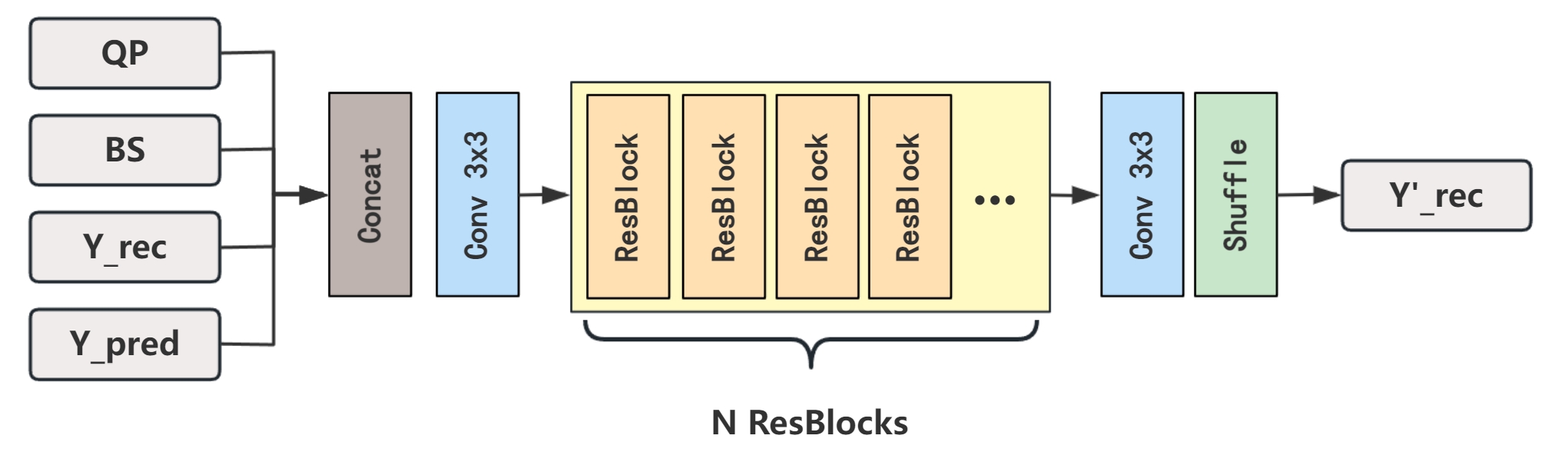} \\
{\small{(a) Network structure for luma component.}} \\[1em]
\epsfig{width=6in,file=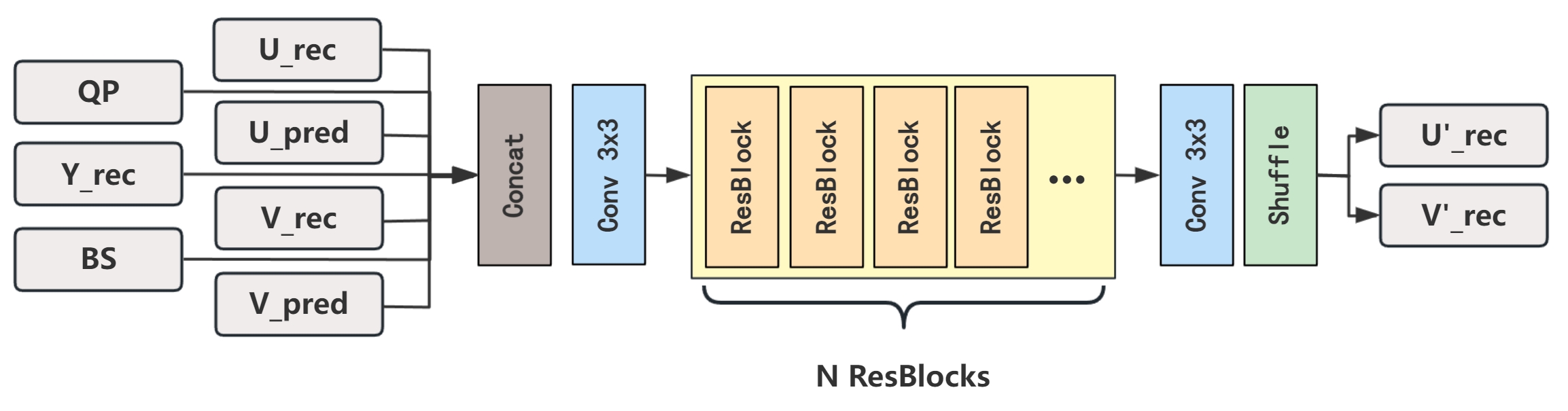} \\
{\small(b) Network structure for chroma components.}
\end{center}
\caption{\label{fig:nnlf}%
Illustration of the network structure regarding the proposed CNN-based in-loop filter.}
\end{figure}

\subsection{Unsymmetric Quaternary Tree}
The challenge of coding unit partitioning has long been a concern in block-based hybrid video coding, as it dictates the dimensions and form of the fundamental coding units. A flexible partitioning framework is instrumental in accurately representing the diverse local content variations. To develop this, we introduce an unsymmetric quaternary tree (UQT) partitioning structure with the objective of enhancing coding efficiency, particularly for larger blocks \cite{zhang2022advanced}. As illustrated in Figure \ref{fig:uqt}, there are four distinct UQT partitioning structures, and these structures denoted as UQT-V1, UQT-V2, UQT-H1, and UQT-H2, are characterized by the location of the largest sub-block, which results in splitting shapes of the areas on the left, right, top, and bottom. In contrast to the conventional QT, UQT divides a block into four sub-parts asymmetrically along a single direction. For the horizontal direction, UQT-H1 and UQT-H2 split the $M \times N$ block into one $M \times N/2$, one $M \times N/4$, and two $M \times N/8$ sub-parts. UQT-V1 and UQT-V2 are along with the vertical direction and divide the $M \times N$ block into one $M/2 \times N$, one $M/4 \times N$, and two $M/8 \times N$ sub-parts. Compared with the QT and Multi-Type Tree (MTT) \cite{bross2021developments}, the smallest sub-blocks generated by UQT could achieve deeper depth with one splitting and capture the rich details more effectively. Furthermore, UQT introduces a novel partitioning pattern that cannot be achieved by QT or MTT, even with identical splitting times. It's worth noting that the dimensions of these sub-blocks are limited to powers of two, obviating the need for introducing new transform shapes.

\subsection{CNN-based In-Loop Filter}
In hybrid video coding, in-loop filters play a vital role in mitigating coding artifacts. These filters encompass the deblocking filter, sample adaptive offset, and adaptive loop filter. They effectively reduce discrepancies between the original input and the reconstructed output. The emergence of convolutional neural networks (CNN) has spurred a deeper exploration of CNN-based in-loop filters, offering the potential to further enhance reconstruction quality. As depicted in Figure \ref{fig:nnlf}, a CNN-based filtering approach with adaptive parameter selection is applied \cite{10.1145/3529107}. To facilitate this improvement, a series of intermediate compression data, acting as supplementary information, is utilized as the input to the network, enriching the prior knowledge. Specifically, prediction signals, boundary strength (BS) computed during the compression process, and the quantization parameter (QP) are integrated as auxiliary information for the network. Furthermore, in the context of the intra-slice filter model, partitioning information is introduced as an additional input. Incorporating the QP as a network input enables the model to achieve a unified framework capable of adapting to various quality levels. An adaptive selection mechanism operates at the slice and block levels, with the model's usage being indicated by signaled flags.

\subsection{Block Importance Mapping} 

The hybrid coding framework minimizes the $R \times \lambda + D$ metric for each block. If a current picture is followed by an identical picture, the coded blocks in the current picture would be re-used identically in the following picture. In this case, $R \times \lambda + 2 \times D$ should be minimized instead. ECM approximates this for RA by setting different QP values for the temporal layers. There is a similarity between the images to be encoded, and this similarity also varies across different sequences and parts of the image. Using a fixed QP for each time layer will result in performance loss, and a separate QP value should be selected for each coding tree unit (CTU) based on the importance of the current image to future images. This importance can be estimated by measuring the difference between motion-compensated pictures. A new algorithm, block importance mapping, was proposed in the proposal \cite{bim} to signal CTU QP delta values in images for reference. The QP value to use for each CTU is based on the estimated importance of a given CTU for future pictures and the QP selected is in the range of $-2$ to $+2$ relative to the picture QP. This tool uses the parameters exported from motion compensated temporal filter (MCTF) to calculate the importance of each CTU and then selects the corresponding delta QP value based on which interval the importance value is located. This method is currently integrated into ECM, but the tool is not enabled by default during encoding. In our scheme, we open it during encoding.

\section*{Experimental Results}

\subsection{Performance with JVET Data Set}
We evaluate the coding performance of the proposed framework with extensive experiments on the JVET dataset compared with ECM10.0. In the evaluation process, random access configuration conforming to the common test condition \cite{vtmtest} is used in the experiments. sequences recommended by JVET are involved in the simulation, including classes A1, A2, B, C, D, E, and F. Due to the testing environment and the coding complexity of the ECM, we did not conduct testing on Class A1 and A2. The QPs are set as 22, 27, 32, 37, and 42. All experiments are conducted on CPU for encoding and decoding testing and neural network inference, and the coding performance is measured by BD-Rate \cite{2001Calculation}, where negative BD-Rate indicates performance improvement. The YUV BD-Rate is calculated by averaging the BD-Rate of Y, U, and V components with weights 6:1:1. Table \ref{lab:UQT} shows the coding performance of the UQT compared with the ECM-10.0 under RA configuration. It can be observed that the UQT achieves 0.17\%, 0.67\%, 0.30\% and 0.25\% BD-Rate savings for Y, U, V and YUV components, respectively on average of class C and D. Table \ref{tab:nnlf} shows that the CNN based in-loop filters can achieve 6.71\%, 14.36\%, 14.87\% and 8.69\% BD-Rate savings for Y, U, V and YUV components, respectively on average of class C and D. When we further enable BIM, our method can bring 6.26\%, 13.33\%, and 12.33\% on the Y, U, and V components compared with ECM-10.0. 

\begin{table}[t]
\centering
\setlength\tabcolsep{3.2pt}
\caption{Performance of the UQT under RA configuration compared with the ECM-10.0}
\label{lab:UQT}
\begin{tabular}{c|ccc|c|c|c}
\hline
\hline
\multirow{2}{*}{\textbf{Class}} &\multicolumn{6}{c}{\textbf{Random Access}} \\
\cline{2-7}
      & \textbf{Y(\%)} & \textbf{U(\%)} & \textbf{V(\%)} & \textbf{YUV(\%)} & \textbf{EncT(\%)} & \textbf{DecT(\%)}\\
\hline
Class C &-0.22\%&-0.35\%&-0.22\%&-0.24\%&123\%&101\%\\   
\hline
Class D  &-0.11\%&-0.99\%&-0.37\%&-0.25\%&118\%&97\%\\
\hline
\textbf{Average} &\textbf{-0.17\%}&\textbf{-0.67\%}&\textbf{-0.30\%}&\textbf{-0.25\%}&\textbf{121\%}&\textbf{99\%}\\
\hline
\hline
\end{tabular}%
\end{table}

\begin{table}[t]
\centering
\setlength\tabcolsep{3.2pt}
\caption{Performance of the CNN-based in-loop filters under RA configuration compared with the ECM-10.0}
\label{tab:nnlf}
\begin{tabular}{c|ccc|c|c|c}
\hline
\hline
\multirow{2}{*}{\textbf{Class}} &\multicolumn{6}{c}{\textbf{Random Access}} \\
\cline{2-7}
      & \textbf{Y(\%)} & \textbf{U(\%)} & \textbf{V(\%)} & \textbf{YUV(\%)} & \textbf{EncT(\%)} & \textbf{DecT(\%)}\\
\hline
Class C &-6.16\%&-13.56\%&-14.19\%&-8.09\%&128\%&26213\%\\   
\hline
Class D  &-7.26\%&-15.15\%&-15.54\%&-9.28\%&122\%&21000\%\\
\hline
\textbf{Average} &\textbf{-6.71\%}&\textbf{-14.36\%}&\textbf{-14.87\%}&\textbf{-8.69\%}&\textbf{125\%}&\textbf{23607\%}\\
\hline
\hline
\end{tabular}%
\end{table}

\begin{table}[t]
\centering
\setlength\tabcolsep{3.2pt}
\caption{Performance of the CNN-based in-loop filters under RA configuration compared with the ECM-10.0}
\label{tab:all}
\begin{tabular}{c|ccc|c|c|c}
\hline
\hline
\multirow{2}{*}{\textbf{Class}} &\multicolumn{6}{c}{\textbf{Random Access}} \\
\cline{2-7}
      & \textbf{Y(\%)} & \textbf{U(\%)} & \textbf{V(\%)} & \textbf{YUV(\%)} & \textbf{EncT(\%)} & \textbf{DecT(\%)}\\
\hline
Class B &-6.51\%&-16.27\%&-14.67\%&-8.75\%&159\%&26138\%\\  
\hline
Class C &-8.00\%&-16.03\%&-15.96\%&-10.00\%&151\%&25075\%\\   
\hline
Class D  &-7.18\%&-17.33\%&-15.98\%&-9.55\%&131\%&15503\%\\
\hline
Class E &-6.51\%&-12.52\%&-12.02\%&-7.95\%&224\%&26750\%\\
\hline
Class F &-3.12\%&-4.52\%&-3.02\%&-3.28\%&172\%&22832\%\\  
\hline
\textbf{Average} &\textbf{-6.26\%}&\textbf{-13.33\%}&\textbf{-12.33\%}&\textbf{-7.91\%}&\textbf{167\%}&\textbf{23260\%}\\
\hline
\hline
\end{tabular}%
\end{table}

\subsection{Performance with CLIC Data Set}
In this section, we evaluate the performance conforming to the test conditions \cite{clic2024} recommended by the CLIC 2024 challenge. In the video compression track, 30 videos are involved. The total decoding time is limited to 20 hours and the decoder size is limited to 4GB. The coding performance is measured by the human rating task under the limited bit rates among all sequences. Due to limitations in GPU resources and decoding time, we only enabled UQT and BIM and did not enable the CNN-based In-Loop Filter in validation phase. The performance in the valid set is shown in Table \ref{valid}. The proposed framework achieves 25.889 dB in terms of PSNR under 0.05 megabits per second (Mbps) of video track. 

\begin{table}[h]
 \centering
  \begin{tabular}{c|cc}
  \hline
  Video Track & 0.5Mbps & 0.05Mbps \\ \hline
  PSNR & - & 25.889 dB \\ \hline
  \end{tabular}
 \caption{Performance of the proposed framework under video compression track in the valid set of CLIC 2024.}
 \label{valid}
\end{table}

\section*{Conclusion}
In this paper, we integrate the UQT and CNN-based in-loop filters into the latest reference software ECM-10.0 and enable the BIM encoding tool. The experimental results show that ECM-10.0, which combines these three coding tools, surpasses the ECM-10.0 with 6.26\%, 13.33\%, and 12.33\% BD-Rate gain. The traditional hybrid coding framework combined with the three coding tools can further improve compression efficiency and has great potential for performance improvement.

\Section{References}
\bibliographystyle{IEEEbib}
\bibliography{refs}

\end{document}